\pdfoutput=1

\documentclass[11pt]{article}

\usepackage[final]{acl}

\usepackage{times}
\usepackage{bm}
\usepackage{amsmath}
\usepackage{latexsym}
\usepackage{booktabs}
\usepackage{multirow}
\usepackage{pifont}
\usepackage{colortbl}
\usepackage{arydshln} 
\usepackage{makecell}

\usepackage[T1]{fontenc}

\usepackage[utf8]{inputenc}

\usepackage{microtype}

\usepackage{inconsolata}

\usepackage{graphicx}

\title{Targeted Distillation for Sentiment Analysis}

\author{
Yice Zhang$^{1}$\thanks{\quad The first two authors contribute equally to this work.}\ , Guangyu Xie$^{1\ast}$, Jingjie Lin$^{1}$,\\ 
\bf Jianzhu Bao$^{4,1}$, Qianlong Wang$^{1}$, Xi Zeng$^{3}$, and Ruifeng Xu$^{1,2}$\thanks{\quad Corresponding Authors}\\
 $^{1}$ Harbin Institute of Technology, Shenzhen, China \\
 $^{2}$ Peng Cheng Laboratory, Shenzhen, China \\
 $^{3}$ The 30th Research Institute of China Electronics Technology Group Corporation \\
 $^{4}$ Nanyang Technological University, Singapore \\
\texttt{zhangyc\_hit@163.com,guangyuxie2001@gmail.com,xuruifeng@hit.edu.cn} \\
}

\begin{document}
\maketitle
\begin{abstract}

This paper explores targeted distillation methods for sentiment analysis\footnote{In this paper, we adopt a broad definition of \textit{sentiment analysis}, which encompasses not only traditional polarity classification but also a range of related tasks such as emotion recognition, irony detection, and stance detection.}, aiming to build compact and practical models that preserve strong and generalizable sentiment analysis capabilities. 
To this end, we conceptually decouple the distillation target into \textit{knowledge} and \textit{alignment} and accordingly propose a two-stage distillation framework. 
Moreover, we introduce \textsc{SentiBench}, a comprehensive and systematic sentiment analysis benchmark that covers a diverse set of tasks across 12 datasets.
We evaluate a wide range of models on this benchmark.
Experimental results show that our approach substantially enhances the performance of compact models across diverse sentiment analysis tasks, and the resulting models demonstrate strong generalization to unseen tasks, showcasing robust competitiveness against existing small-scale models.\footnote{We release our code, data, and model weights at \url{https://github.com/HITSZ-HLT/Sentiment-Distillation}.}
\end{abstract}

\section{Introduction}

Sentiment analysis, aiming to identify and extract subjective information from user-generated content \cite{liu2021sentiment}, has emerged as a significant research area in natural language processing, garnering widespread attention \cite{zhang2018deep,wankhade2022survey}.
Recent studies demonstrate that large language models (LLMs) exhibit remarkable capabilities and achieve state-of-the-art performance in diverse sentiment analysis tasks \cite{zhang-etal-2024-sentiment,wang2024is,smid-etal-2024-llama}. Despite these advancements, the practical application of LLMs faces significant challenges. Deploying these models incurs considerable computational costs, and fine-tuning them for enhanced task-specific performance demands greater computational resources.

\begin{figure}[t]
\centering
\includegraphics[width=0.92\linewidth]{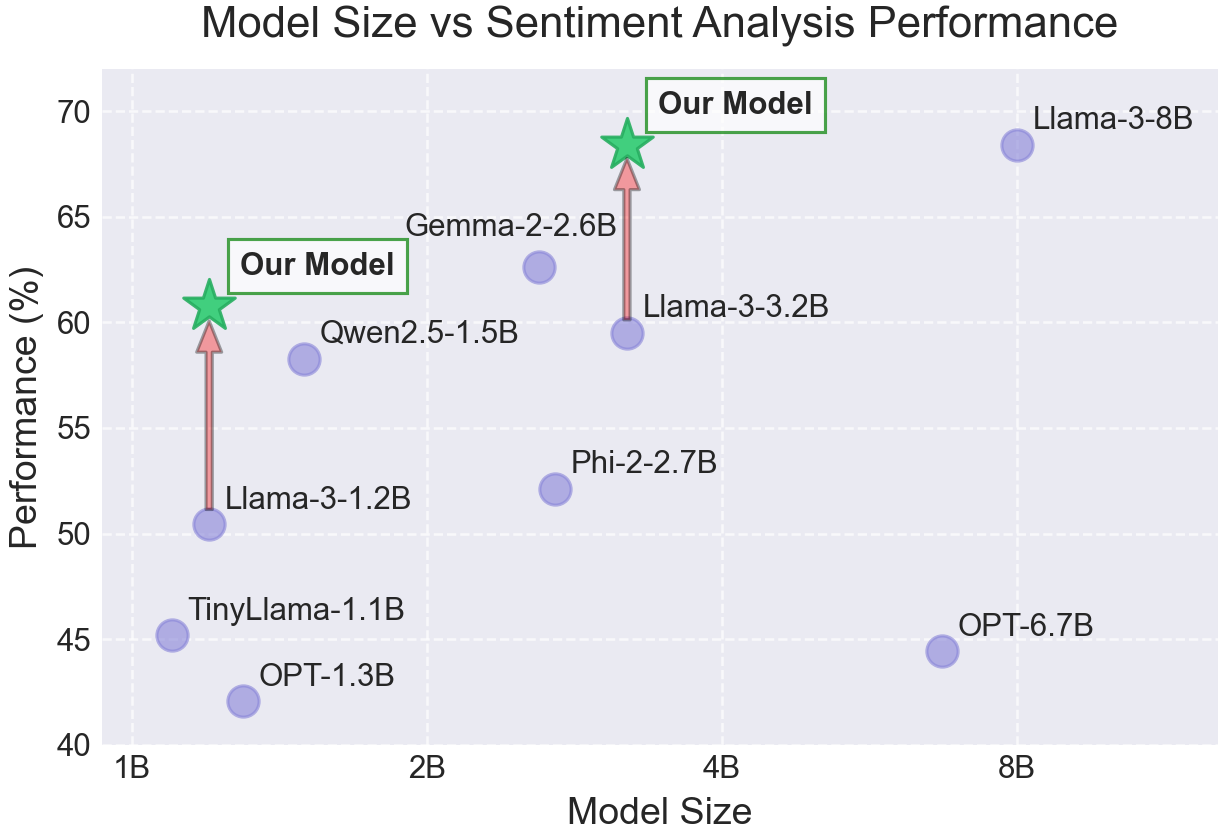}
\caption{
The comparison of our distilled model with other small-scale models in terms of the
average performance on \textsc{SentiBench} 
($F_1$-score, \%). 
}
\label{fig:intro}
\end{figure}

To reduce computational overhead, researchers are increasingly turning to knowledge distillation techniques \cite{hinton2015distillingknowledgeneuralnetwork}.
These works focus on transferring general capabilities from advanced LLMs to their more cost-efficient counterparts through carefully curated instructions \cite{alpaca,vicuna2023,wu-etal-2024-lamini}.
However, when substantial size gaps exist between teacher and student models, such generic distillation is challenging due to the difficulty in developing instructions with sufficient diversity and scale.
Consequently, students often merely mimic the output style of teacher LLMs while performing poorly on specialized downstream tasks \cite{gudibande2023falsepromiseimitatingproprietary}.
In contrast, existing works demonstrate that for a specific application class, LLMs can be 
potentially 
approximated by a much smaller model \cite{xu-etal-2023-inheritsumm,kim2024prometheus,zhou2024universalner}. This suggests that targeted distillation towards specialized capabilities offers a more practical and promising direction.

Motivated by these insights, 
this paper explores targeted distillation specifically for sentiment analysis.
We conceptually decouple the distillation target into \textit{knowledge} and \textit{alignment} and propose a two-stage distillation framework.
The {first} stage, termed \textbf{knowledge-driven distillation} (\textsc{KnowDist}), focuses on transferring fundamental sentiment analysis capabilities, thereby improving the student model's potential performance.
In \textsc{KnowDist}, we devise a multi-perspective prompting strategy to elicit comprehensive sentiment-related knowledge from the teacher LLM and systematically transfer this knowledge to the student model.
The {second} stage, termed \textbf{in-context learning distillation} (\textsc{ICLDist}), transfers prompt-following capabilities in sentiment analysis to optimize the student model's task alignment.
In \textsc{ICLDist}, we enable the student model to follow task-specific instructions and demonstrations by mimicking the teacher LLM's responses on few-shot samples.
When constructing few-shot samples, we implement format and task diversification strategies to strengthen the generalization of \textsc{ICLDist}.

To facilitate a systematic evaluation, we develop \textsc{SentiBench}, a comprehensive sentiment analysis benchmark, comprising 3 task categories across 12 datasets.
Our extensive experimentation on this benchmark reveals several key findings: (1) Our approach demonstrates substantial advantages over generic distillation methods, achieving effective distillation of LLMs' sentiment analysis capabilities. Specifically, the student model achieves a 10\% improvement in the average $F_1$-score across various tasks, with a particularly remarkable gain of 38\% in irony detection. (2) Our approach enables the 1.2B model to outperform the original 3.2B model, and the 3.2B model to surpass the original 8B model. As illustrated in Figure~\ref{fig:intro}, the resulting models exhibit strong competitiveness against other small-scale models.  (3) Further analysis reveals the complementary nature of \textsc{KnowDist} and \textsc{ICLDist} and validates the effectiveness of each component in our approach.

\begin{figure*}[h]
\centering
\includegraphics[width=0.85\linewidth]{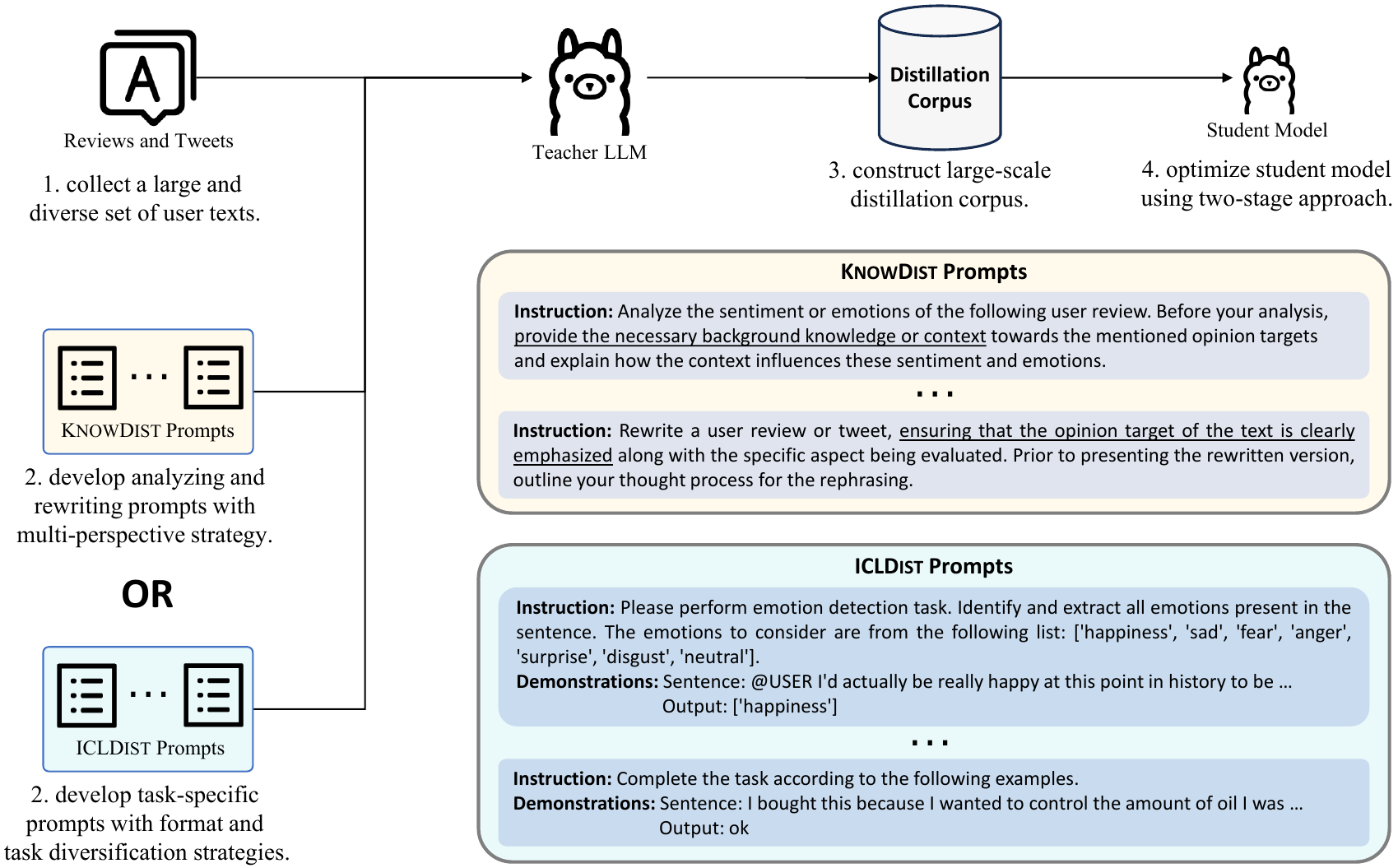}
\caption{
Illustration of our distillation process, consisting of four steps: data collection, prompt construction, corpus generation, and student model optimization.
}
\label{fig:framework}
\end{figure*}

\section{Two-stage Distillation Framework}

Following \citet{alpaca,vicuna2023,wu-etal-2024-lamini}, we distill the capabilities of LLMs by making the student model learn from the teacher LLM's output $\bm y$ for specific prompts.
Our prompts are composed of instructions $\bm i$, demonstrations $\bm d$ (which may be empty), and input texts $\bm x$. This process can be formulated as follows:
\begin{align}
\bm y &= {\cal M}(\bm i, \bm d, \bm x; \theta_{T}),
\end{align}
\begin{align}
\hat{\theta}_S &= \underset{\theta_S}{\operatorname{argmax}} \hspace{-5pt}\sum_{\bm i,\bm d,\bm x,\bm y}\hspace{-5pt}\log P_{\cal M}(\bm y\ |\ \bm i, \bm d, \bm x; \theta_S),
\end{align}
where $\cal M$ denotes the teacher or student model, and $\theta_T$ and $\theta_S$ denote their respective parameters.

In contrast to prior research, this paper focuses on distilling the LLMs' capability specifically for sentiment analysis. Prior to distillation, we decouple the target into \textit{sentiment-related knowledge} and \textit{task alignment}. (1) The knowledge reflects a model's ability to comprehend the sentiments expressed in text, including accurate interpretation of sentiment expressions, precise targeting, and possession of the requisite background knowledge. The capacity of this knowledge within the model shapes its potential performance in sentiment analysis tasks.
(2) The alignment refers to the model's ability to follow task-specific instructions and demonstrations, \textit{i.e.}, its in-context learning ability.
Such alignment capability determines the model's observable performance in sentiment analysis tasks.
Based on this decoupling, we develop a distillation framework consisting of two stages: knowledge-driven distillation (\textsc{KnowDist}) and in-context learning distillation (\textsc{ICLDist}).

\subsection{Knowledge-Driven Distillation}

At this stage, we develop two distinct prompting methods to elicit sentiment-related knowledge from LLMs. The first directs LLMs to \textit{analyze} the sentiments embedded within the given text, while the second instructs LLMs to \textit{rewrite} the text while maintaining its original sentiment. Crucially, both methods require LLMs to provide their reasoning process before generating the final output.

To enhance the effectiveness of these prompting methods, we devise a \textbf{multi-perspective prompting strategy}.
This strategy defines four different perspectives: (1) \textsc{Expression}: centering on subjective words and phrases during analyzing or rewriting; 
(2) \textsc{Target}: focusing on the specific entities and their associated aspects being evaluated; 
(3) \textsc{Emotion}: highlighting the emotional states and psychological reactions expressed in the text; (4) \textsc{Background}: incorporating contextual information and domain knowledge necessary for understanding the sentiment. This strategy guides the analyzing and rewriting process from these four perspectives, thereby eliciting a more comprehensive range of sentiment-related knowledge.
The specific prompts can be found in Appendix \ref{app:a}.

We employ these prompting methods to perform \textsc{KnowDist}, as illustrated in Figure \ref{fig:framework}. Firstly, we collect a large and diverse set of user-generated content, including movie, product, and restaurant reviews, and tweets. Secondly, we construct various analyzing and rewriting prompts following our multi-perspective prompting strategy. Thirdly, we apply these prompts to guide the teacher LLM in interpreting existing sentiments within these texts and actively exploring and generating diverse sentiment expression patterns. This process yields a large-scale corpus enriched with sentiment-related knowledge. Finally, we leverage this corpus to optimize the student model, thereby enhancing its fundamental sentiment analysis capabilities.

\subsection{In-Context Learning Distillation}

After the \textsc{KnowDist} stage, we optimize the student model's alignment in specific sentiment analysis tasks. To achieve this, we construct task-specific prompts comprising instructions, demonstrations, and input text. We then train the student model to mimic the teacher LLM's output on these prompts, aiming to enhance its ability to follow task-specific instructions and demonstrations. However, this method faces a major challenge: we cannot anticipate all potential downstream tasks, making it impossible to prepare corresponding prompts in the \textsc{ICLDist} stage. Consequently, the student model may underperform on previously unseen tasks.
For example, when using sentiment classification and emotion recognition as distillation tasks, the student model performs poorly on unseen tasks such as irony detection.

\definecolor{lightgray}{gray}{0.95}

\begin{table*}[ht]
\centering
\fontsize{8.5pt}{0.84\baselineskip}\selectfont
\begin{tabular}{l c ccc c ccc c ccc c ccc c c} 
\toprule
\textbf{Task} & \textbf{Dataset} & \textbf{Train} & \textbf{Dev} & \textbf{Test} & \textbf{\#Class} & \textbf{Metric}\\
\midrule 
\noalign{\smallskip}
\multicolumn{7}{c}{\textsc{Basic Sentiment Analysis}} \\
\hdashline[2pt/4pt]
& IMDb  & 3000 & 300 & 1000 & 2 & macro\_f1 \\
        \multirow{-2}{*}{{Document-level sentiment classification}}
& Yelp2 & 3000 & 300 & 1000 & 2 & macro\_f1 \\
    \rowcolor{lightgray}
& SST2  & 3000 & 300 & 1821 & 2 & macro\_f1\\
    \rowcolor{lightgray}
        \multirow{-2}{*}{{Sentence-level sentiment classification}}
& Twitter17 & 3000 & 300 & 1000 & 3 & macro\_f1\\
\midrule
    \noalign{\smallskip}
\multicolumn{7}{c}{\textsc{Multifaceted Sentiment Analysis}} \\
\hdashline[2pt/4pt]
Irony detection & Irony18 & 3000 & 300 & 784 & 2 & macro\_f1 \\
    \rowcolor{lightgray}
Emotion recognition & Emotion20 & 3000 & 300 & 1421 & 4 & macro\_f1\\
Stance detection & P-Stance & 3000 & 300 & 2157 & 3 & macro\_f1 \\
    \rowcolor{lightgray}
Intimacy analysis & \textsc{Mint}-English & 1287 & 300 & 396 & 3 & macro\_f1 \\
\midrule
    \noalign{\smallskip}
\multicolumn{7}{c}{\textsc{Fine-Grained Sentiment Analysis}} \\
\hdashline[2pt/4pt]
Aspect term sentiment analysis & Rest16 & 1600 & 400 & 676 & - & micro\_f1 \\
    \rowcolor{lightgray}
Aspect category sentiment analysis & Rest16 & 1600 & 400 & 676 & - & micro\_f1 \\
Aspect sentiment quad prediction
& Rest16 & 1264 & 316 & 544 & - & micro\_f1 \\
    \rowcolor{lightgray}
Structured sentiment analysis
& Opener & 1744 & 249 & 499 & - &  sentiment\_graph\_f1\\
\bottomrule
\end{tabular}
\caption{
Task overview and dataset statistics in \textsc{SentiBench}.
We perform downsampling on some datasets to ensure computational efficiency. For sampling details, please refer to Appendix \ref{app:b.1}.
}
\label{tab:sentibench}
\end{table*}

To enhance generalization on unseen tasks, we maximize the diversity of the distillation prompts, introducing \textbf{format and task diversification strategies}.
Format diversification refers to using varied prompt formats for the same task to mitigate overfitting. We devise three specific strategies to achieve this.
The first is to alter {label word} formats, replacing standard labels like \texttt{positive/negative/neutral} with alternatives like \texttt{good/bad/ok} or \texttt{+1/-1/0}.
The second is to diversify {label taxonomies}, for the emotion recognition task, employing various classification systems, such as Ekman's taxonomy \cite{10.1037/0033-295x.99.3.550} or the GoEmotions taxonomy \cite{demszky-etal-2020-goemotions}.
The third is to utilize minimized instructions, placing task information within demonstrations, exemplified by prompts like ``\textit{Complete the task according to the following examples}''.

Task diversification refers to incorporating a variety of tasks other than sentiment analysis during the \textsc{ICLDist} stage. To this end, we select about 100 natural language understanding tasks from the \textsc{Super-NaturalInstructions} dataset \cite{wang-etal-2022-super} and construct corresponding prompts. We intentionally exclude sentiment analysis tasks from this selection to prevent overlap with downstream evaluation tasks. While these tasks are not directly related to sentiment analysis, we hypothesize that they can enhance 
the model's general prompt-following capability.

The \textsc{ICLDist} process is illustrated in Figure \ref{fig:framework}. Similar to knowledge collection, we first gather a large volume of user-generated content. Next, we select sentiment classification and emotion recognition as distillation tasks\footnote{For these two tasks, we construct a collection of 55 demos. Specifically, we first randomly select user-generated text of a suitable length, then use GPT-4o to annotate its sentiment and emotion labels, and finally perform manual verification to ensure the quality of the annotations.} and construct prompts by randomly applying our format diversification strategies. Additionally, we incorporate the task diversification strategy to generate supplementary prompts. 
We then collect the teacher LLM's responses to these prompts, resulting in a task-alignment corpus.
Finally, we optimize the student model on this corpus to enhance its task alignment.

\section{\textsc{SentiBench}}

To systematically assess LLMs' sentiment analysis capabilities, we develop a comprehensive benchmark. This benchmark encompasses three typical categories: basic sentiment analysis, multifaceted sentiment analysis, and fine-grained sentiment analysis. Multifaceted and fine-grained analyses extend the breadth and depth of evaluation, respectively. For each category, we carefully curate representative tasks and their corresponding datasets. Table \ref{tab:sentibench} provides a comprehensive overview of these tasks along with detailed dataset statistics.

\vspace{5pt}
\noindent
\textbf{Basic sentiment analysis (BSA)} aims to classify the overall sentiment polarity expressed in texts.
We collect and curate four widely-adopted sentiment classification datasets, covering both document and sentence levels.
For document-level sentiment classification, we incorporate IMDb \cite{maas-etal-2011-learning} and Yelp2 \cite{NIPS2015_250cf8b5}, while for sentence-level classification, we utilize SST2 \cite{socher-etal-2013-recursive} and Twitter17 \cite{rosenthal-etal-2017-semeval}.

\vspace{5pt}
\noindent
\textbf{Multifaceted sentiment analysis (MSA)} extends beyond merely identifying sentiment polarity, focusing instead on recognizing a broader range of human emotional states \cite{zhang-etal-2024-sentiment}.
Our benchmark incorporate four MSA tasks:
(1) Irony detection identifies instances whether the intended meaning 
contradicts the literal expression;
(2) Emotion recognition categorizes text into discrete emotional categories, such as anger, joy, sadness, and optimism;
(3) Stance detection determines the position or attitude towards a specific target or topic;
(4) Intimacy analysis assesses the degree of interpersonal closeness reflected in the text, examining the model's understanding of social information.
For these tasks, we curate the following datasets:
Irony18 \cite{van-hee-etal-2018-semeval} for irony detection,
Emotion20 \cite{mohammad-etal-2018-semeval,barbieri-etal-2020-tweeteval} for emotion recognition,
P-Stance \cite{li-etal-2021-p} for stance detection,
and \textsc{Mint}-English \cite{pei-etal-2023-semeval} for intimacy analysis.

\vspace{5pt}
\noindent
\textbf{Fine-grained sentiment analysis (FSA)} transcends basic sentiment analysis, aiming to recognize a spectrum of sentiment elements, thereby providing a more complete picture of opinions.
Our benchmark incorporates four FSA tasks:
(1) Aspect term sentiment analysis (ATSA) extracts aspect terms from the text and determining their sentiment polarities;
(2) Aspect category sentiment analysis (ACSA) identifies the evaluated aspect categories and their sentiment polarities;
(3) Aspect sentiment quad prediction (ASQP) structures opinions into fine-grained quadruples comprising category, aspect, opinion, and polarity;
(4) Structured sentiment analysis (SSA) formalizes opinions as quadruples containing a sentiment holder, target, expression, and polarity.
For these tasks, we curate the following datasets:
Rest16 \cite{pontiki-etal-2016-semeval,zhang-etal-2021-aspect-sentiment} 
for ATSA, ACSA, and ASQP,
and Opener 
\cite{barnes-etal-2022-semeval} for SSA.

Our benchmark is partially inspired by \citet{zhang-etal-2024-sentiment}. Our work differs from theirs in the following aspects:
(1) We develop a reorganized evaluation task taxonomy;
(2) Following the revised taxonomy, we refine the tasks and datasets;
(3) We conduct comprehensive evaluations across a range of LLMs, with a particular attention to small-scale models.

\newcommand{\sixf}[1]{\fontsize{7pt}{\baselineskip}\selectfont(+#1)}
\newcommand{\sixd}[1]{\fontsize{7pt}{\baselineskip}\selectfont(-#1)}

\begin{table*}[ht]
\centering
\fontsize{8.5pt}{0.84\baselineskip}\selectfont
\setlength\tabcolsep{0.54pt}
\begin{tabular}{
p{94pt}
>{\centering\arraybackslash}m{25pt}
>{\centering\arraybackslash}m{25pt}
>{\centering\arraybackslash}m{25pt}
>{\centering\arraybackslash}m{25pt}
>{\centering\arraybackslash}m{25pt}
>{\centering\arraybackslash}m{25pt}
>{\centering\arraybackslash}m{25pt}
>{\centering\arraybackslash}m{25pt}
>{\centering\arraybackslash}m{25pt}
>{\centering\arraybackslash}m{25pt}
>{\centering\arraybackslash}m{25pt}
>{\centering\arraybackslash}m{25pt}
>{\centering\arraybackslash}m{45pt}} 
\toprule
\multirow{2}*{\textbf{Models}} & \multicolumn{4}{c}{\textbf{BSA}} & \multicolumn{4}{c}{\textbf{MSA}} & \multicolumn{4}{c}{\textbf{FSA}} & \multirow{2}*{\textbf{Avg}}\\
\cmidrule(lr){2-5}  \cmidrule(lr){6-9} \cmidrule(lr){10-13} 
& {IMDb} & {Yelp2} & {SST2} 
& {Twitter} & {Irony} & {Emoti.}  
& {Stance} & {Intim.} & {ATSA} & {ACSA} & {ASQP} & SSA \\
\midrule
Llama-3-8B  & 94.17	& 98.07 & 95.90 & 66.58 & 82.63	& 73.00 & 75.86 & 49.85 & 54.41 & 64.57 & 19.67 & 31.91 & 67.22\\
Llama-3-70B & 95.30 & 98.10 & 97.14 & 68.75 & 83.99 & 75.87 & 85.21 & 53.68 & 63.78 & 75.21 & 31.03 & 45.29 & 72.78\\
GPT-3.5 & 93.70 & 98.30 & 96.31 & 60.15 & 78.64 & 75.61 & 79.99 & 52.63 & 56.43 & 66.67 & 30.30 & 44.01 & 69.40 \\
\midrule
OPT-1.3B & 78.94 & 91.37 & 77.10 & 39.32 & 51.18 & 43.98 & 53.93 & 32.65 & 11.39 & 19.06 & 1.72 & 3.92 & 42.05\\
TinyLlama-1.1B & 71.27 & 84.13 & 78.01 & 34.21 & 56.15 & 50.05 & 57.25 & 36.95 & 26.76 & 29.42 & 4.24 & 13.68 & 45.18 \\
Phi-2-2.7B & 87.03 & 96.10 & 90.63 & 59.59 & 47.52 & 45.53 & 55.36 & 31.61 & 39.71 & 46.54 & 9.60 & 16.31 & 52.13 \\
Gemma-2-2.6B & 92.39 & 97.40 & 94.17 & 56.02 & 70.68 & 68.85 & 73.99 & 42.57 & 48.00 & 50.27 & 18.03 & 39.08 & 62.62\\

\noalign{\smallskip}
\hdashline[2pt/4pt]
\noalign{\smallskip}
Llama-3-1.2B & 87.65 & 94.80 & 88.93 & 58.78 & 35.80 & 58.07 & 60.78 & 25.60 & 33.80 & 36.09 & 8.05 & 16.91 & 50.44 \\
~+ Distill. \textit{w/} Alpaca-data & 89.13	& 94.37 & 91.08 & 58.02 & 33.01 & 60.24 & 64.02 & 26.10 & 36.18 & 37.71 & 8.72	& 16.44 & 51.25\sixf{0.81} \\
~+ Distill. \textit{w/} Lamini-data 
& 89.26 & 94.63 & 91.14 & 62.90 &	38.05 & 50.61 & 63.92 & 27.90 &	35.03 & 41.89 & 8.30 & 18.80 & 51.87\sixf{1.43} \\
~+ \textsc{Know} \& \textsc{ICLDist}
& 93.07 & 97.70 & 94.53 & 68.37 & 73.80 & 76.79 & 69.94 & 35.39 & 39.01 & 47.82 & 11.69 & 21.18 & 60.77\sixf{10.33} \\

\noalign{\smallskip}
\hdashline[2pt/4pt]
\noalign{\smallskip}

Qwen-2.5-1.5B & 91.92 & 97.30 & 92.33 & 52.39 & 65.80 & 63.61 & 70.90 & 35.73 & 37.66 & 53.25 & 18.47 & 20.08 & 58.29 \\

~+ Distill. \textit{w/} Alpaca-data 
& 92.07 & 96.63 & 92.25 & 51.84 & 66.24 & 54.76 & 70.31 & 28.20 & 41.50 & 57.15 & 18.76 & 18.72 & 57.37\sixd{0.92} \\

~+ Distill. \textit{w/} Lamini-data 
& 92.80 & 97.60 & 93.08 & 57.75 & 71.94 & 51.37 & 71.54 & 29.10 & 40.74 & 57.87 & 18.81 & 15.27 & 58.16\sixd{0.13} \\

~+ \textsc{Know} \& \textsc{ICLDist} & 93.80 & 98.10 & 95.99 & 65.89 & 71.69 & 72.57 & 74.83 & 47.31 & 51.36 & 53.98 & 22.93 & 19.70 & 64.01\sixf{5.72} \\

\noalign{\smallskip}
\hdashline[2pt/4pt]
\noalign{\smallskip}
Llama-3-3.2B  & 92.57 & 96.53 & 93.59 & 61.45 & 64.00 & 68.88 & 71.43 & 33.32 & 46.37 & 51.66 & 11.09 & 23.10 & 59.50\\
~+ Distill. \textit{w/} Alpaca-data & 92.37 & 97.37 & 93.92 & 57.70 & 66.59 & 64.47 & 72.05 & 28.70 & 44.70 & 50.77 & 14.19 & 24.63 & 58.96\sixd{0.54}\\
~+ Distill. \textit{w/} Lamini-data & 92.80 & 97.33 & 94.91 & 62.07 & 70.10 & 65.61 & 72.49 & 40.28 & 50.29 & 52.62 & 16.06 & 27.36 & 61.83\sixf{2.33}\\
~+ \textsc{Know} \& \textsc{ICLDist}
& \textbf{94.30} & \textbf{98.17} & \textbf{95.41} & \textbf{69.57} & \textbf{85.25} & \textbf{77.47} & \textbf{75.10} & \textbf{48.24} & \textbf{53.24} & \textbf{66.01} & \textbf{22.95} & \textbf{35.16} & \textbf{68.41\sixf{8.91}} \\
\bottomrule
\end{tabular}
\caption{
Experimental results on \textsc{SentiBench} ($F_1$-score, \%).
}
\label{tab:main-results}
\end{table*}

\section{Experiments}
\subsection{Experimental Setup}

\textbf{Implementation Details.} 
The teacher LLM is set to Llama-3.1-70B-Instruct \cite{grattafiori2024llama3herdmodels}, while Llama-3.2-1.2B-Instruct, Qwen-2.5-1.5B-Instruct\footnote{\url{https://qwenlm.github.io/blog/qwen2.5/}}, and Llama-3.2-3.2B-Instruct are employed as student models.
For distillation, we curate a large and diverse corpus of user-generated texts from IMDb \cite{NguyenWASSA2014long}, Yelp\footnote{\url{https://www.yelp.com/dataset}}, Amazon\footnote{\url{https://nijianmo.github.io/amazon/index.html}}, and Twitter\footnote{\url{https://archive.org/details/twitterstream}}. 
We preprocess this corpus by decontaminating it for the downstream datasets and eliminating duplicates using \texttt{simhash}.
We then apply the proposed prompting methods to these user texts and obtain 1M \textsc{KnowDist} samples and 400K \textsc{ICLDist} samples.
We further supplement the \textsc{ICLDist} corpus with 100K general task samples from the \textsc{super-naturalInstruction} \cite{wang-etal-2022-super} dataset.
The 1.2B and 1.5B models are optimized using the complete training set, while the 3.2B model is trained on a subset containing 200K \textsc{KnowDist} samples and 100K \textsc{ICLDist} samples. 
The hyperparameter settings are provided in Appendix \ref{app:c.1}.

After distillation, we evaluate the student model on \textsc{SentiBench} using in-context learning, with dataset statistics shown in Table \ref{tab:sentibench}. The specific prompts for each task are detailed in Appendix \ref{app:b.2}. During evaluation, we randomly sample 4 examples from the dev set as demonstrations. To ensure generation stability, we set the temperature parameter to 0 during model inference. To mitigate the impact of randomness, we conduct each evaluation using 3 different random seeds and report the average results.

\vspace{5pt}
\noindent
\textbf{Baselines.}
We compare our approach with generic distillation methods. Specifically, we train the student model using existing instruction-following datasets, including the 52K data constructed by \citet{alpaca} (alpaca-data), and the 2.58M data developed by \citet{wu-etal-2024-lamini} (lamini-data).
Besides, we evaluate a diverse set of models for reference: 
(1) Llama-3 series models, spanning different scales (8B and 70B variants); (2) several small-scale models ranging from 1B to 3B parameters, including OPT-1.3B \cite{zhang2022optopenpretrainedtransformer}, TinyLlama-1.1B-Chat-v1.0 \cite{zhang2024tinyllamaopensourcesmalllanguage}, Phi-2-2.7B\footnote{\url{https://huggingface.co/microsoft/phi-2}}, and Gemma-2-2.6B-it \cite{gemma_2024}; and (3) GPT-3.5\footnote{Available at \url{https://chat.openai.com/}. The specific model used is \texttt{gpt-3.5-turbo-0125}.}.

\subsection{Main Results}

Table \ref{tab:main-results} presents the comparison results on \textsc{SentiBench}. We observe that two generic distillation methods yield only marginal and unstable gains in sentiment analysis performance, with the student model showing average $F_1$-score improvements below 2.33\%. These limited improvements suggest that utilizing generic distillation methods to transfer sentiment analysis capabilities is ineffective. In contrast, our approach, namely \textsc{Know} \& \textsc{ICLDist}, significantly enhances the sentiment analysis performance of the student model. Specifically, our approach achieves an average improvement of 10.33\%, 5.72\%, and 8.91\%. The most striking improvement is observed in irony detection of Llama-3.2-1.2B-Instruct, where the $F_1$-score increases dramatically from 35.80\% to 73.80\% - an improvement of 38.00\%. These results demonstrate the effectiveness of our approach in transferring sentiment analysis capabilities from the LLM to its more efficient counterparts.

Furthermore, the experimental results in Table \ref{tab:main-results} reveal several additional insights. Firstly, within the Llama-3 family, we observe a clear positive correlation between model size and performance, with Llama-3-70B achieving the best results, surpassing GPT-3.5.
Secondly, our approach empowers the 1.2B model to outperform the original 3.2B model, and the 3.2B model to surpass the original 8B model. Moreover, the distilled models demonstrate strong competitive performance compared to other small-scale models and GPT-3.5, also illustrated in Figure \ref{fig:intro}. 
Thirdly, our approach demonstrates consistent performance gains across different model families, including Llama-3 and Qwen2.5, highlighting its broad generalization.
Fourthly, with Llama-3-70B as the teacher LLM, our approach enables Llama-3-3.2B to achieve comparable performance to the teacher on sentiment classification, irony detection, and emotion recognition.
Finally, both the distilled models and other small-scale models show inferior performance on intimacy analysis and tuple extraction tasks (\textit{i.e.,} ASQP and SSA). These tasks require a deep understanding of social context and advanced structured extraction capabilities, presenting promising directions for future research.

\begin{figure}[t]
\centering
\includegraphics[width=.99\linewidth]{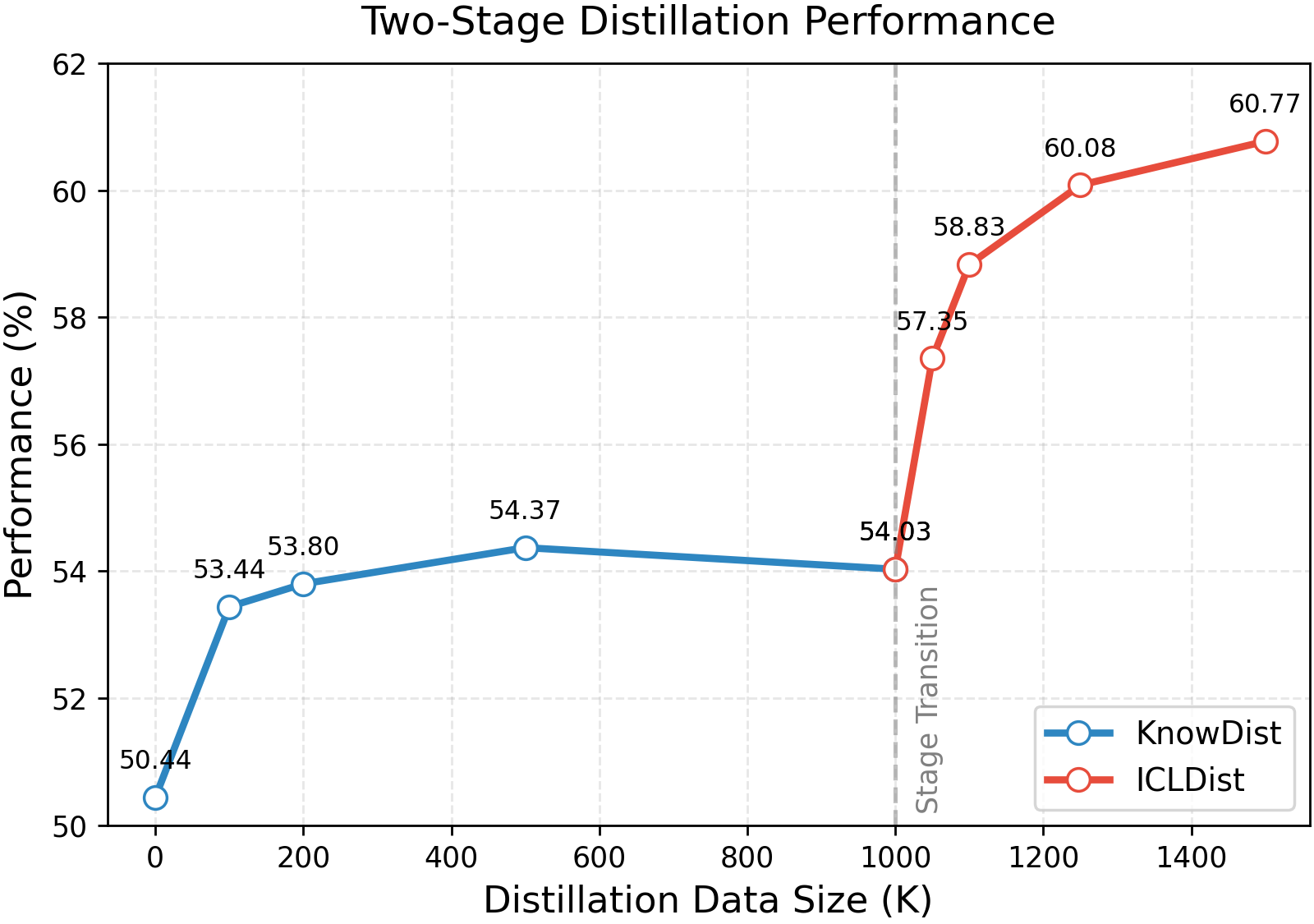}
\caption{
Performance trend of the student model with varying volumes of distillation data (\%).
Here, performance refers to the average $F_1$-score on \textsc{SentiBench}.
}
\label{fig:data-scale}
\end{figure}

\subsection{Analyis of Two-stage Optimization}

Our distillation framework consists of two stages: \textsc{KnowDist} and \textsc{ICLDist}. 
Below, we conduct an in-depth analysis of these two stages, aiming to distinguish their respective roles and investigate how they complement each other.

Figure \ref{fig:data-scale} illustrates the performance trends of the student model across different volumes of distillation data. We observe that in both stages, model performance generally improves as the data volume increases. Moreover, the improvements brought by \textsc{ICLDist} are notably more pronounced and efficient. These observations raise two natural questions: (1) Given \textsc{ICLDist}'s superior performance, is the \textsc{KnowDist} stage essential to the framework? (2) Could we simplify the framework by merging data from both stages into a unified optimization process?

For the first question, we conduct fine-tuning experiments using the training samples from \textsc{SentiBench}. The results in Table \ref{tab:finetuned} demonstrate that under the fine-tuning setting, both \textsc{KnowDist} and \textsc{ICLDist} can enhance the student model's sentiment analysis performance. Notably, \textsc{KnowDist} achieves more substantial improvements, 
which contrasts with the in-context learning results in 
Figure \ref{fig:data-scale}.
These findings support our claims: \textsc{KnowDist} strengthens the student model's fundamental sentiment analysis capabilities, while \textsc{ICLDist} optimizes task alignment. When sufficient labeled samples are available for downstream task alignment, the benefits of \textsc{ICLDist}'s task alignment become less significant. However, such labeled data is often scarce in real-world applications. Consequently, both \textsc{KnowDist} and \textsc{ICLDist} are essential components of our framework.

\begin{table}[ht]
\centering
\fontsize{8.5pt}{0.83\baselineskip}\selectfont
\begin{tabular}{lcccc ccc c ccc c ccc c c} 
\toprule
 Models & MSA & FSA \\
\midrule
Llama-3-1.2B & 73.61 & 68.78 \\

~+~\textsc{KnowDist} & {76.12\sixf{2.51}} & {69.70\sixf{0.92}} \\
~+~\textsc{ICLDist} & 74.64\sixf{1.03} & 69.30\sixf{0.52}\\

\bottomrule
\end{tabular}
\caption{
Experimental results on MSA and FSA categories under fine-tuning settings ($F_1$-score, \%). Models are fine-tuned jointly on all tasks within each category.
}
\label{tab:finetuned}
\end{table}

\begin{table}[b]
\centering
\fontsize{8.5pt}{0.83\baselineskip}\selectfont
\setlength\tabcolsep{5.2pt}
\begin{tabular}{lcccc ccc c ccc c ccc c c} 
\toprule
 Models & BSA & MSA & FSA \\
\midrule
Llama-3-1.2B & 82.54 & 45.06 & 23.72 \\

~+~\textsc{KnowDist} & 83.65\sixf{1.01} & 50.65\sixf{5.59} & 27.11\sixf{3.39} \\
~+~\textsc{ICLDist} & 87.83\sixf{5.29} & 58.75\sixf{13.69} & 27.55\sixf{3.83} \\

~+~\textsc{Unified} & 87.21\sixf{4.67} & 53.57\sixf{8.51} & 27.45\sixf{3.73} \\
~+~\textsc{Two-stage} & 88.06\sixf{5.52} & 60.70\sixf{15.64} & 27.74\sixf{4.02} \\
\bottomrule
\end{tabular}
\caption{
Comparison results between unified optimization and two-stage optimization ($F_1$-score, \%).
}
\label{tab:unified-vs-two-stage}
\end{table}

For the second question, we conduct experiments to compare unified optimization against two-stage optimization, with results presented in Table \ref{tab:unified-vs-two-stage}. The results reveal that unified optimization not only significantly underperforms two-stage optimization but also falls behind using \textsc{ICLDist} alone. This suggests that unified optimization would disrupt the distillation process and impair the learning efficiency of the student model. These findings demonstrate the necessity of two-stage optimization in our framework.

\subsection{Ablation Studies}

\noindent
\textbf{\textsc{KnowDist.}}
In this stage, we employ two distinct prompting methods (analyzing and rewriting) to elicit sentiment-related knowledge from the teacher LLM and introduce a multi-perspective prompting (MPP) strategy to enhance their effectiveness. As shown in Table \ref{tab:ablation-knowdist}, the MPP strategy significantly improves the performance of both prompting methods. Specifically, for the analyzing method, MPP 
yields additional improvements of 3.90\% and 1.46\% on MSA and FSA, respectively.
Among the two prompting methods, the analyzing method achieves more substantial performance gains, while the combination of both methods leads to better overall performance. These results demonstrate the effectiveness of each sub-component within \textsc{KnowDist}.

\begin{table}[h]
\centering
\fontsize{8.5pt}{0.83\baselineskip}\selectfont
\setlength\tabcolsep{2pt}
\begin{tabular}{c c ccc c ccc c ccc c ccc c c} 
\toprule
\textsc{Dist} & \textsc{Anl} & \textsc{Rw} & MPP & BSA & MSA & FSA  \\
\midrule
\ding{55} & - & - & - & 82.54 & 45.06 & 23.72\\
\ding{51} & \ding{51} & \ding{55} & \ding{55} & 83.69\sixf{1.15} & 45.72\sixf{0.66} & 26.07\sixf{2.35}  \\
\ding{51} & \ding{51} & \ding{55} & \ding{51} & {83.92\sixf{1.38}} & 49.62\sixf{4.56} & {27.53\sixf{3.81}} \\
\ding{51} & \ding{55} & \ding{51} & \ding{55} & 83.44\sixf{0.90} & 44.98\sixd{-0.08} & 24.85\sixf{1.13}  \\
\ding{51} & \ding{55} & \ding{51} & \ding{51} & 82.77\sixf{0.23} & 47.90\sixf{2.84} & 26.02\sixf{2.30} \\
\ding{51} & \ding{51} & \ding{51} & \ding{51} & 83.65\sixf{1.11} & {50.65\sixf{5.59}} & 27.11\sixf{3.39} \\
\bottomrule
\end{tabular}
\caption{
Ablation results of \textsc{KnowDist} ($F_1$-score, \%). \textsc{Anl} and \textsc{Rw} denote analyzing and rewriting respectively, and MPP stands for multi-perspective prompting. 
The distillation samples used are 200K.
}
\label{tab:ablation-knowdist}
\end{table}

\begin{table}[htb]
\centering
\fontsize{8.5pt}{0.83\baselineskip}\selectfont
\setlength\tabcolsep{5.1pt}
\begin{tabular}{ccccc ccc c ccc c ccc c c} 
\toprule
 \textsc{Dist} & LW & LT & MI & TD & Seen & Unseen\\
\midrule

 \ding{55} & - & - & - & - & 77.65 & 31.00 \\
 \ding{51} & \ding{55} & \ding{55} & \ding{55} & \ding{55} & 85.36\sixf{7.71} & 33.53\sixf{2.53} \\
 \ding{51} & \ding{51} & \ding{55} & \ding{55} & \ding{55} & 85.18\sixf{7.53} & 33.91\sixf{2.91}\\
 \ding{51} & \ding{51} & \ding{51} & \ding{55} & \ding{55} & 85.44\sixf{7.79} & 34.07\sixf{3.07} \\
 \ding{51} & \ding{51} & \ding{51} & \ding{51} & \ding{55} & 85.08\sixf{7.43} & 35.09\sixf{4.09} \\
 \ding{51} & \ding{55} & \ding{55} & \ding{55} & \ding{51} & 85.64\sixf{7.99} & 37.52\sixf{6.52} \\
 \ding{51} & \ding{51} & \ding{51} & \ding{51} & \ding{51} & 85.01\sixf{7.36} & 38.79\sixf{7.79} \\

\bottomrule
\end{tabular}
\caption{
Ablation results of \textsc{ICLDist} ($F_1$-score, \%). LW, LT, and MI denote the format diversification of Label Words, Label Taxonomies, and Minimized Instructions respectively, while TD represents Task Diversification. We divide tasks in \textsc{SentiBench} into seen and unseen categories during distillation, where seen tasks include sentiment classification and emotion recognition, while the rest are considered unseen.
The distillation samples used are 100K.
}
\label{tab:ablation-icldist}
\end{table}

\noindent
\textbf{\textsc{ICLDist.}}
A key challenge in this stage is the limited generalization to tasks unseen during distillation. To address this challenge, we develop several diversification strategies. As shown in Table \ref{tab:ablation-icldist}, without these strategies, the performance improvement on unseen tasks (2.53\%) is substantially lower than that on seen tasks (7.71\%), confirming our concerns about generalization. After incorporating our diversification strategies, the student model achieves a significant performance gain on unseen tasks (7.79\%), reaching a comparable level of improvement to seen tasks. These results demonstrate the effectiveness of our diversification strategies in enhancing model generalization.

\subsection{Discussions}

\textbf{Effect of Teacher LLMs.}
We experiment with different teacher LLMs in our distillation framework to analyze their impact. The results in Table \ref{tab:effect-of-teacher} reveal that teacher quality significantly influences distillation effectiveness, as larger teacher LLMs generally lead to more substantial improvements. Furthermore, we make two noteworthy observations. 
First, even when using identical models for both teacher and student, distillation has the potential to enhance the student's sentiment analysis performance.
This result suggests the potential for leveraging distillation to achieve self-improvement in specialized domains. 
Second, larger teachers do not always lead to better performance, as evidenced in FSA tasks, where the 8B teacher slightly outperforms the 70B teacher. We hypothesize that larger teachers may sometimes pose greater learning challenges for smaller student models, warranting further exploration in future work.

\begin{table}[ht]
\centering
\fontsize{8.5pt}{0.83\baselineskip}\selectfont

\begin{tabular}{l c ccc c ccc c ccc c ccc c c} 
\toprule
{Teachers} & BSA & MSA & FSA\\
\midrule
No Distill. & 82.54 & 45.06 & 23.72\\
Llama-3-1.2B & 80.45\sixd{2.09} & 46.33\sixf{1.27} & 22.53\sixd{1.19}\\
Llama-3-3.2B & 85.85\sixf{3.31} & 51.05\sixf{5.99} & 27.59\sixf{3.87}\\
Llama-3-8B  & 85.90\sixf{3.36} & 57.16\sixf{12.10} & 29.02\sixf{5.30} \\ 
Llama-3-70B & 88.06\sixf{5.52} & 60.70\sixf{15.64} & 27.74\sixf{4.02} \\
\bottomrule
\end{tabular}
\caption{
Experimental results using different teacher LLMs in our distillation framework ($F_1$-score, \%).
}
\label{tab:effect-of-teacher}
\end{table}

\begin{table}[t]
\centering
\fontsize{8.5pt}{0.83\baselineskip}\selectfont
\setlength\tabcolsep{2.23pt}
\begin{tabular}{l c ccc c ccc c ccc c ccc c c} 
\toprule

{Models} & Human. & Social. & STEM & Other & Avg\\
\midrule

Llama-3-1.2B & 42.87 & 51.16 & 39.68 & 52.11 & 46.12 \\
~+~\textsc{Know} \& \textsc{ICLDist} & 43.14 & 52.62 & 40.17 & 53.40 & 46.94\\
\bottomrule
\end{tabular}
\caption{
Experimental results on 5-shot MMLU (accuracy, \%).
Our evaluation is conducted using LM-Evaluation-Harness provided at
\url{https://github.com/meta-llama/llama-cookbook}.
}
 
\label{tab:mmlu-results}
\end{table}

\vspace{5pt}
\noindent
\textbf{Results on MMLU.}
A potential concern of targeted distillation towards specialized capabilities is the possible degradation of the model's general abilities. To investigate this concern, we conduct evaluations on the Massive Multitask Language Understanding (MMLU) benchmark \cite{hendrycks2021measuring}. As shown in Table \ref{tab:mmlu-results}, we find that our distillation approach not only avoids any deterioration but also results in a slight improvement. This indicates that our distillation approach can enhance specialized capabilities without compromising general capabilities.

\section{Related Work}

\textbf{Applying LLMs for Sentiment Analysis.} Many researchers adopt in-context learning methods to harness LLMs for sentiment analysis tasks \cite{zhang-etal-2024-sentiment,wang2024is,bai-etal-2024-compound,xu2023limitschatgptextractingaspectcategoryopinionsentiment}. To enhance the effectiveness of in-context learning, research has branched into (1) selecting semantically relevant examples for demonstrations \cite{wang-etal-2024-context,xu-etal-2024-improving,10.1145/3626772.3657932}, (2) utilizing chain-of-thought reasoning to enhance sentiment inference \cite{fei-etal-2023-reasoning}, and (3) integrating relevant background knowledge to generate more nuanced and informed predictions \cite{10.1145/3604237.3626866}. Furthermore, a range of studies explore fine-tuning methods to better align LLMs with sentiment analysis tasks \cite{fatemi2023comparativeanalysisfinetunedllms,smid-etal-2024-llama,simmering2023largelanguagemodelsaspectbased}.

\vspace{5pt}
\noindent 
\textbf{Knowledge Distillation from LLMs.}
In light of the high computational demands or issues of proprietary access, many studies explore knowledge distillation techniques \cite{hinton2015distillingknowledgeneuralnetwork} to transfer the capabilities of LLMs into more compact and accessible models  \cite{alpaca,vicuna2023,wu-etal-2024-lamini,chen2024knowledgedistillationblackboxlarge,muralidharan2024compact}.
Recent advancements in this field concentrate on optimizing distillation objectives to improve the efficiency and effectiveness of the distillation process \cite{zhong-etal-2024-revisiting,gu2024minillm,ko2024distillm,agarwal2024onpolicy}.
Besides, there is a growing trend towards distilling specialized capabilities from LLMs,
including leveraging LLMs as annotators to generate pseudo-labeled data \cite{ding-etal-2023-gpt,xu-etal-2023-inheritsumm,kim2024prometheus,zhou2024universalner,he2024annollmmakinglargelanguage} and synthesizing task-specific data from scratch \cite{ye-etal-2022-zerogen,he-etal-2023-targeted,gao2023selfguided,xu2024ds2absadualstreamdatasynthesis}.

\section{Conclusions}

This paper explores targeted distillation for sentiment analysis, introducing a two-stage distillation framework. The first stage (\textsc{KnowDist}) aims to transfer fundamental sentiment analysis capabilities, while the second stage (\textsc{ICLDist}) focuses on transfering task-specific prompt-following abilities. Besides, we develop a comprehensive and systematic benchmark for sentiment analysis, named \textsc{SentiBench}. Extensive experiments on this benchmark demonstrate that our framework enables the 1.2B model to outperform the original 3.2B model, and the 3.2B model to outperform the original 8B model, 
showing strong competitiveness compared to other small-scale models.

\section*{Limitations}

We list the potential limitations of this paper: \begin{itemize} 
\item Our approach transfers knowledge directly from teacher LLMs without filtering or processing their responses. This direct transfer may propagate erroneous or low-quality information to the student model, potentially impacting its performance. Future work could explore quality control mechanisms during the distillation process. 
\item As shown in Table \ref{tab:main-results}, our model exhibits unsatisfactory performance on tuple extraction tasks (\textit{i.e.}, ASQP and SSA). This suggests the need for specialized optimization of structured extraction capabilities. \end{itemize} 
We believe that these limitations offer promising directions for future research.

\section*{Ethics Statement}
Large language models for sentiment analysis have enabled progress in areas such as public health and commercial applications; yet their reliance on large-scale pretraining corpora raises ethical concerns, including risks of privacy violations, cultural and annotator subjectivity, and systematic harms to marginalized groups \cite{ehics-sheet-for-sa}. While knowledge distillation substantially improves efficiency and deployability, prior work shows that it can also transfer and intensify existing biases, exacerbating disparities across sentiment classes and demographic subgroups. 

Accordingly, ethical evaluation of distilled sentiment models should not only emphasize improvements in overall performance but also recognize the risks of propagating biases and exacerbating disparities across categories and social subgroups \cite{distillationfairness}. Therefore, the community should place greater emphasis on assessing subgroup- and category-level fairness, accompanied by clearer documentation of risks and limitations. In addition, exploring fairness-aware distillation methods and developing practical guidelines could help mitigate potential misuse in sensitive or high-stakes applications.

\section*{Acknowledgments}
This work was supported by the National Natural Science Foundation of China 62176076 and 62576120,  Natural Science Foundation of Guang Dong 2023A1515012922, the Major Key Project of PCL2023A09,  CIPSC-SMP-ZHIPU Large Model Cross-Disciplinary Fund ZPCG20241119405 and Key Laboratory of Computing Power Network and Information Security, and Ministry of Education under Grant No.2024ZD020.

\bibliographystyle{acl_natbib}

\appendix

\newpage

\section*{Organization of Appendices}
We structure the appendix into four sections: \begin{itemize} \item Appendix \ref{app:a} details the complete prompts utilized in our distillation framework; \item Appendix \ref{app:b} provides the construction details and evaluation prompts of \textsc{SentiBench}; \item Appendix \ref{app:c} outlines the hyperparameter settings of the two-stage optimization and the computational cost incurred during the construction of the distillation corpus; \item Appendix \ref{app:d} provides additional experimental results, which include the evaluation of teacher quality, comparison with reasoning-enhanced methods, and the case study of the distilled model. \end{itemize}

\section{Distillation Prompts}
\label{app:a}

\subsection{Prompts in Knowledge-Driven Distillation}

In this stage,
we develop two distinct prompting methods (analyzing and rewriting) along with a multi-perspective prompting strategy. The corresponding prompts for these methods are presented in Tables \ref{tab:analyzing-prompt} and \ref{tab:rewriting-prompt}.

\subsection{Prompts in In-Context Learning Distillation}

In this stage, we employ sentiment classification and emotion recognition as distillation tasks and devise multiple strategies to enhance prompt diversity, including label word (LW) diversification, label taxonomies (LT) diversification, and minimized instruction (MI) strategies. Tables \ref{tab:sc-prompt} and \ref{tab:er-prompt} present the specific prompts. In practice, these prompts contain a random number of demonstrations ranging from 1 to 16. These tables only show examples with one demonstration.

\begin{table}[t]
\centering
\fontsize{8.5pt}{0.82\baselineskip}\selectfont
\setlength\tabcolsep{1pt}
\begin{tabular}{p{7.5cm}} 
\toprule
\textbf{Analyzing - \textsc{Basic}}\\

Analyze the overall sentiment of the following text. Provide a brief explanation supporting your conclusion.

Text: \textcolor{blue}{\{Text\}}\\

\midrule
\textbf{Analyzing - \textsc{Target}}\\
Given a text, list the mentioned opinion targets, analyzing the evaluated aspects and the corresponding sentiments. Provide brief explanations supporting your conclusions.

Text: \textcolor{blue}{\{Text\}}\\

\midrule
\textbf{Analyzing - \textsc{Expression}}\\

Identify all sentiment expressions in the following text, i.e., those words or phrases that convey sentiment or emotion. For each sentiment expression, provide a clear explanation of how it contributes to the overall sentiment.

Text: \textcolor{blue}{\{Text\}}\\

\midrule
\textbf{Analyzing - \textsc{Emotion}}\\

Analyze the following text and identify any emotions being expressed, such as happiness, sadness, anger, fear, surprise, or disgust. For each emotion identified, explain how it is reflected in the text.

Text: \textcolor{blue}{\{Text\}}\\

\midrule
\textbf{Analyzing - \textsc{Background}}\\

Analyze the sentiment or emotions of the following text. Before your analysis, provide the necessary background knowledge or context towards the mentioned opinion targets and explain how the context influences these sentiment and emotions.

Text: \textcolor{blue}{\{Text\}}\\

\bottomrule
\end{tabular}
\caption{
Analyzing prompts in \textsc{KnowDist}.
}

\label{tab:analyzing-prompt}
\end{table}

\begin{table}[t]
\centering
\fontsize{8.5pt}{0.82\baselineskip}\selectfont
\setlength\tabcolsep{1pt}

\begin{tabular}{p{7.5cm}} 
\toprule
\textbf{Rewriting - \textsc{Basic}}\\

Rewrite the following text to ensure it retains the original sentiment and tone, but presents it in a rephrased or alternative way. Prior to presenting the rewritten version, outline your thought process for the rephrasing.

Text: \textcolor{blue}{\{Text\}}\\

\midrule
\textbf{Rewriting - \textsc{Target}}\\

Rewrite the following text, ensuring that the opinion target of the text is clearly emphasized along with the specific aspect being evaluated. Prior to presenting the rewritten version, outline your thought process for the rephrasing.

Text: \textcolor{blue}{\{Text\}}\\

\midrule
\textbf{Rewriting - \textsc{Expression}}\\

Rewrite the following text while focusing on the sentiment expressions used. Prior to presenting the rewritten version, outline your thought process for the rephrasing.

Text: \textcolor{blue}{\{Text\}}\\

\midrule
\textbf{Rewriting - \textsc{Emotion}}\\

Rewrite the following text by highlighting the expressed emotions (such as happiness, sadness, anger, fear, surprise, or disgust). Prior to presenting the rewritten version, outline your thought process for the rephrasing.

Text: \textcolor{blue}{\{Text\}}\\

\midrule
\textbf{Rewriting - \textsc{Background}}\\

Rewrite the following text to enhance sentiment clarity by incorporating necessary background knowledge or context. Prior to presenting the rewritten version, outline your thought process for the rephrasing.

Text: \textcolor{blue}{\{Text\}}\\

\bottomrule
\end{tabular}
\caption{
Rewriting prompts in \textsc{KnowDist}.
}

\label{tab:rewriting-prompt}
\end{table}

\begin{table}[!t]
\centering
\fontsize{8.5pt}{0.82\baselineskip}\selectfont
\setlength\tabcolsep{1pt}

\begin{tabular}{p{7.5cm}} 
\toprule
\textbf{Sentiment Classification - \textsc{Basic}}\\

Please perform sentiment classification task. The label should be one of the following: [`positive', `negative', `neutral']. In your classification, consider the overall content, tone, emotional language, and any contextual clues that indicate the sentiment behind the sentence. Do not provide any reasoning or explanation and directly output the final answer.\\
\\

Sentence: I bought this because I wanted to control the amount of oil I was using. I read the other reviews and the ... \\
Output: neutral\\
\\
Sentence: A fabulous social commentary is illustrated between the lines that you can enjoy privately in your mind while ... \\
Output:\\

\midrule
\textbf{Sentiment Classification - \textsc{LW}}\\
Please perform sentiment classification task. The label should be one of the following: [`+1', `-1', `0']/[`POS', `NEG', `NEU']/[`good', `bad', `ok']. In your classification, consider the overall content, tone, emotional language, and any contextual clues that indicate the sentiment behind the sentence. Do not provide any reasoning or explanation and directly output the final answer.\\
\\

Sentence: I bought this because I wanted to control the amount of oil I was using. I read the other reviews and the ...
Output: 0\\
\\
Sentence: If your planting several rows of garden veggies, ie: corn beans, etc, this is a great time saver. You must make ...\\
Output: \\

\midrule
\textbf{Sentiment Classification - \textsc{MI}}\\

Please complete the task according to the following examples. Do not provide any reasoning or explanation and directly output the final answer.\\
\\
Sentence: I couldn't use this cable. But it is not the fault of the cable. I ordered it to use with my new kodak printer. I ... \\
Output: neutral\\
\\
Sentence: This is a good family game, easy to learn, and straightforward to play. Also helpful in teaching US geography ...\\

Output: \\

\bottomrule
\end{tabular}
\caption{
Sentiment classification prompts in \textsc{ICLDist}.
}
\label{tab:sc-prompt}
\end{table}

\begin{table}[!t]
\centering
\fontsize{8.8pt}{0.8\baselineskip}\selectfont
\setlength\tabcolsep{1pt}
\begin{tabular}{p{7.5cm}} 
\toprule
\textbf{Emotion Recognition - \textsc{Basic}}\\

Please perform emotion detection task. Identify and extract all emotions present in the sentence. The emotions to consider are from the following list: [`happiness', `sad', `fear', `anger', `surprise', `disgust', `neutral']. In your analysis, take into account the language used, context, and any emotional expressions or cues that indicate multiple emotions. Do not provide any reasoning or explanation and directly output the final answer.\\
\\
Sentence: I just received a pair 38x30 VIP and they were a bit loose around the waste, and the legging was long enough ...\\
Output: [`disgust', `neutral', `sadness']\\
\\
Sentence: First, the title is misleading. One might expect a book called \"Stumbling on happiness\" to perhaps provide ...\\

Output: \\

\midrule
\textbf{Emotion Recognition - \textsc{LT}}\\
Please perform emotion detection task. Identify and extract all emotions present in the sentence. The emotions to consider are from the following list: [`neutral', `curiosity', `confusion', `amusement', `gratitude', `admiration', `pride', `approval', `realization', `surprise', `excitement', `joy', `relief', `caring', `optimism', `desire', `love', `fear', `nervousness', `grief', `sadness', `remorse', `disapproval', `disappointment', `anger', `annoyance', `embarrassment', `disgust']. In your analysis, take into account the language used, context, and any emotional expressions or cues that indicate multiple emotions. Do not provide any reasoning or explanation and directly output the final answer.\\
\\

Sentence: Let me start by saying that I have read as many Agatha Christie books as I possibly could. Sad Cypress ...\\

Output: [`curiosity', `admiration', `surprise', `disappointment', `disapproval']\\
\\
Sentence: I put this in my Garage and the humidity that comes out of the end is good for the wood in this kind of ...\\
Output: \\

\midrule
\textbf{Emotion Recognition - \textsc{MI}}\\

Please complete the task according to the following examples. Do not provide any reasoning or explanation and directly output the final answer.\\
\\

Sentence: I really don't get how this game got such good ratings. My only guess is that people just like game of ...\\
Output: [`disgust', `neutral', `anger']\\
\\
Sentence: This wonderful allegory is highly entertaining for a young person and deeply inspiring for an adult who is ...\\
Output: \\

\bottomrule
\end{tabular}
\caption{
Emotion recognition prompts in \textsc{ICLDist}.
}
\label{tab:er-prompt}
\end{table}

\section{\textsc{SentiBench} Details}
\label{app:b}

In contrast to the taxonomy proposed by \citet{zhang-etal-2024-sentiment}, we introduce a more coherent and practically comprehensive task taxonomy with three categories: basic sentiment analysis (BSA), multifaceted sentiment analysis (MSA), and fine-grained sentiment analysis (FSA). For BSA, we refine the sentiment classification category in \citet{zhang-etal-2024-sentiment} by excluding aspect-level tasks, as they conceptually belong to fine-grained analysis rather than basic sentiment classification, thereby creating cleaner categorical boundaries. 
We further expand MSA by incorporating intimacy analysis, which requires models to capture subtle social dynamics and interpersonal affect. 
Finally, our FSA category extends fine-grained analysis beyond traditional aspect-term sentiment tasks by incorporating aspect category sentiment analysis (ACSA) and structural sentiment analysis (SSA), allowing a more complete assessment of models’ ability to capture compositional and context-dependent sentiment.

\subsection{Datasets}
\label{app:b.1}

For computational efficiency, we sample from the original datasets. Specifically:
\begin{itemize}
    \item For basic sentiment analysis tasks, we randomly sample 3000 instances from each training set of IMDb, Yelp2, SST2, and Twitter17. For validation, we randomly sample 300 instances from each validation set of these datasets. For testing, we randomly sample 1,000 instances each from the test sets of IMDb, Yelp2, and Twitter17, while retaining the original test set for SST2 due to its smaller size.
    \item For multifaceted sentiment analysis tasks, we randomly sample 3000 instances each from the training sets of Irony18, Emotion20, and P-Stance. For validation, we randomly sample 300 instances from each validation set of these four datasets. Due to their limited sizes, we retained all original test sets for these tasks.
    \item For fine-grained sentiment analysis tasks, we retain all original datasets due to their limited sizes.
\end{itemize}

\begin{table}[!t]
\centering
\fontsize{8.5pt}{0.82\baselineskip}\selectfont
\setlength\tabcolsep{1pt}
\begin{tabular}{p{7.5cm}} 
\toprule
\textbf{BSA - IMDb}\\

Please perform Sentiment Analysis task. Given the sentence, assign a sentiment polarity label from [`negative', `positive']. Return label only without any other text.\\

\\
Sentence: I have to agree with MR. Caruso Jr Lanza,s was the finest voice god had to offer if only he could have ...\\

Label: positive\\
\\
Sentence: I watched this film with a bunch of friends at a Halloween party last night. I got to say that the ... \\

Label:\\

\midrule

\textbf{BSA - Yelp2}\\
Please perform Sentiment Analysis task. Given the sentence, assign a sentiment polarity label from [`negative', `positive']. Return label only without any other text.\\

\\
Sentence: I'm so glad Yelp has added verbal descriptions for the star system as, "Meh. I've experienced better." ...\\
Label: negative\\
\\
Sentence: We went here yesterday for lunch, it wasnt packed at all and the lunch prices are good. They start you off ...\\

Label:\\

\midrule
\textbf{BSA - SST2}\\

Please perform Sentiment Analysis task. Given the sentence, assign a sentiment polarity label from [`negative', `positive']. Return label only without any other text.\\
\\
Sentence: as relationships shift , director robert j. siegel allows the characters to inhabit their world without ...\\

Label: positive\\
\\
Sentence: this is one of polanski 's best films .\\

Label:\\

\midrule

\textbf{BSA - Twitter17}\\

Please perform Sentiment Analysis task. Given the sentence, assign a sentiment polarity label from [`negative', `positive', `neutral']. Return label only without any other text.\\
\\

Sentence: "It's 4.33am, I can't sleep. Just bought two pairs of sun glasses online n caught up on Hulk Hogan news ...\\

Label: positive\\
\\
Sentence: @user Bull vs Corbin is the gold standard for bad no DQ matches, this was a close second.\\

Label:\\

\bottomrule
\end{tabular}
\caption{
The prompts for basic sentiment analysis (BSA) task.
}
\label{tab:bsa-prompt}
\end{table}

\begin{table}[!t]
\centering
\fontsize{8.5pt}{0.82\baselineskip}\selectfont
\setlength\tabcolsep{1pt}
\begin{tabular}{p{7.5cm}} 
\toprule
\textbf{MSA - Irony Detection - Irony18}\\

Please perform Irony Detection task. Given the sentence, assign a sentiment label from [`irony', `non-irony']. Return label only without any other text.\\
\\

Sentence: @user I infer that you are besmirching coffee, but that can't be right

Label: non-irony\\
\\

Sentence: Just walked in to \#Starbucks and asked for a "tall blonde" Hahahaha

Label:\\

\midrule
\textbf{MSA - Emotion Recognition - Emotion20}\\
Please perform Emotion Detection task. Given the sentence, assign a emotion label from [`anger', `joy', `sadness', `optimism']. Return the label only without any other text.\\
\\

Sentence: it's pretty depressing when u hit pan on ur favourite highlighter

Label: sadness\\

\\
Sentence: @user Interesting choice of words... Are you confirming that governments fund \#terrorism? Bit of an open door, but still...

Label:\\

\midrule
\textbf{MSA - Stance Detection - P-Stance}\\

Please perform Stance Detection task. Given the sentence, assign a sentiment label expressed by the author towards "Bernie Sanders" from [`against', `favor']. Return label only without any other text.\\

\\

Sentence: ? seriously - no hate but what leadership . dude is loosing sensibility and MIA. Bernie though has ...\\

Label: favor (opinion towards `Bernie Sanders')\\

\\
Sentence: He's the ONLY ONE Where have I heard that before? No, Bernie is NOT the only one The Democrats ...\\

Label:\\

\midrule
\textbf{MSA - Intimacy Analysis - \textsc{Mint}-English}\\

Please perform Intimacy Detection task. Given the sentence, assign an intimacy label from [`not intimate', `moderately intimate', `highly intimate']. Return label only without any other text.\\
\\

Sentence: Would God be pleased if you were working to hasten the apocalypse?

Label: not intimate\\

\\
Sentence: @tessavirtue Happy new year!!!! Love u

Label:\\

\bottomrule
\end{tabular}
\caption{
The prompts for multifaceted sentiment analysis (MSA) task.
}
\label{tab:msa-prompt}
\end{table}

\begin{table*}[htbp]
\centering
\fontsize{8.5pt}{0.82\baselineskip}\selectfont
\setlength\tabcolsep{1pt}
\begin{tabular}{p{15.5cm}} 
\toprule
\textbf{FSA - ATSA - Rest16}\\

Please perform Aspect Term Sentiment Analysis task. Given the sentence, extract all (aspect term, sentiment polarity) pairs.\\
\\

Sentence: I had the best ravioli ever.\\
Label: [(`ravioli', `positive')]\\
\\
Sentence: Green Tea creme brulee is a must!\\
Label:\\

\midrule
\textbf{FSA - ACSA - Rest16}\\
Please perform aspect-level sentiment analysis task. Given the sentence, tag all (aspect category, sentiment) pairs. Aspect category should be selected from [`ambience general', `drinks prices', `drinks quality', `drinks style\_options', `food prices', `food quality', `food style\_options', `location general', `restaurant general', `restaurant miscellaneous', `restaurant prices', `service general'], and sentiment should be selected from [`negative', `neutral', `positive']. If there are no target-sentiment pairs, return an empty list. Otherwise return a python list of tuples containing two strings in double quotes. Please return python list only, without any other comments or texts.\\
\\
Sentence: I pray it stays open forever.

Label: [(`restaurant general', `positive')]\\
\\

Sentence: Serves really good sushi.\\

Label:\\

\midrule
\textbf{FSA - ASQP - Rest16}\\

Please perform Aspect Sentiment Quad Prediction task. Given the sentence, extract all (aspect term, aspect category, opinion, sentiment polarity) quadruples.\\

1. Aspect category should be selected from [`ambience general', `drinks prices', `drinks quality', `drinks style\_options', `food general', `food prices', `food quality', `food style\_options', `location general', `restaurant general', `restaurant miscellaneous', `restaurant prices', `service general'].\\

2. Sentiment polarity should be selected from [`negative', `neutral', `positive'].\\

3. If there is no aspect term, use `NULL' as the aspect term. Only aspect term can be `NULL', aspect category, opinion and sentiment polarity CANNOT be `NULL'.

4. Please return python list only, without any other comments or texts.\\
\\

Sentence: Make sure you try this place as often as you can .\\
Label: [(`restaurant general', `place', `try', `positive')]\\
\\
Sentence: All their menu items are a hit , and they serve mimosas\\
Label:\\

\midrule
\textbf{FSA - SSA - Opener}\\

Please perform the Structured Sentiment Analysis task. Given a sentence, extract all opinion tuples in the format (holder, target, sentiment expression, sentiment polarity). 

Each tuple should contain:

- Holder: The entity expressing the sentiment, if there is no explicit holder, use `NULL' as the holder.

- Target: The entity being evaluated, if there is no explicit target, use `NULL' as the target.

- Sentiment Expression: The phrase conveying the sentiment.

- Sentiment Polarity: The polarity of the sentiment, either positive, negative, or neutral.

Follow these rules:

1. If there is no sentiment expression, return `NULL' for all fields.

2. Please return python list only, without any other comments or texts.\\
\\

Sentence: A beautiful wellness hotel\\
Label: [(`NULL', `wellness hotel', `beautiful', `positive']\\
\\

Sentence: We went foor a cheap city trip and that 's what we have got .\\
Label:\\

\bottomrule
\end{tabular}
\caption{
The prompts for fine-grained sentiment analysis (FSA) task.
}
\label{tab:fsa-prompt}
\end{table*}

\subsection{Task Prompts}
\label{app:b.2}

The corresponding prompts for BSA, MSA, FSA tasks are presented in Tables \ref{tab:bsa-prompt}, \ref{tab:msa-prompt}, and \ref{tab:fsa-prompt}.

\begin{table}[!t]
\centering
\fontsize{8.5pt}{0.82\baselineskip}\selectfont
\begin{tabular}{lc} 
\toprule
\textbf{Hyper-parameter} & \textbf{Value} \\ 
\midrule
Batch Size & 128 \\ 
Learning Rate & 5e-6 \\ 
Training Epoch & 4 \\ 
Learning Rate Deacy & Cosine \\ 
Warmup Step Ratio & 0.01 \\ 
Weight Decay & 0.1 \\ 
Adam $\beta_1$ & 0.9 \\ 
Adam $\beta_2$ & 0.95 \\ 
\bottomrule
\end{tabular}
\caption{
Hyperparameters for \textsc{KnowDist}'s optimization for Llama-3.2-1.2B-Instruct.
}
\label{tab:know-hp-llama-1b}
\end{table}

\begin{table}[!t]
\centering
\fontsize{8.5pt}{0.82\baselineskip}\selectfont
\begin{tabular}{lc} 
\toprule
\textbf{Hyper-parameter} & \textbf{Value} \\ 
\midrule
Batch Size & 128 \\ 
Learning Rate & 1e-5 \\ 
Training Epoch & 4 \\ 
Learning Rate Deacy & Linear \\ 
Warmup Step Ratio & 0.02 \\ 
Weight Decay & 0.01 \\ 
Adam $\beta_1$ & 0.9 \\ 
Adam $\beta_2$ & 0.999 \\ 
\bottomrule
\end{tabular}
\caption{
Hyperparameters for \textsc{ICLDist}'s optimization for Llama-3.2-1.2B-Instruct.
}
\label{tab:icl-hp-llama-1b}
\end{table}

\begin{table}[!t]
\centering
\fontsize{8.5pt}{0.82\baselineskip}\selectfont
\begin{tabular}{lc} 
\toprule
\textbf{Hyper-parameter} & \textbf{Value} \\ 
\midrule
Batch Size & 128 \\ 
Learning Rate & 5e-5 \\ 
Training Epoch & 4 \\ 
Learning Rate Deacy & Cosine \\ 
Warmup Step Ratio & 0 \\ 
Weight Decay & 0.1 \\ 
Adam $\beta_1$ & 0.9 \\ 
Adam $\beta_2$ & 0.999 \\ 
\bottomrule
\end{tabular}
\caption{
Hyperparameters for \textsc{KnowDist}'s optimization for Qwen-2.5-1.5B-Instruct.
}
\label{tab:know-hp-qwen-1.5b}
\end{table}

\begin{table}[!t]
\centering
\fontsize{8.5pt}{0.82\baselineskip}\selectfont
\begin{tabular}{lc} 
\toprule
\textbf{Hyper-parameter} & \textbf{Value} \\ 
\midrule
Batch Size & 128 \\ 
Learning Rate & 3e-5 \\ 
Training Epoch & 4 \\ 
Learning Rate Deacy & Cosine \\ 
Warmup Step Ratio & 0 \\ 
Weight Decay & 0.1 \\ 
Adam $\beta_1$ & 0.9 \\ 
Adam $\beta_2$ & 0.999 \\ 
\bottomrule
\end{tabular}
\caption{
Hyperparameters for \textsc{ICLDist}'s optimization for Qwen-2.5-1.5B-Instruct.
}
\label{tab:icl-hp-qwen-1.5b}
\end{table}

\begin{table}[!t]
\centering
\fontsize{8.5pt}{0.82\baselineskip}\selectfont
\begin{tabular}{lc} 
\toprule
\textbf{Hyper-parameter} & \textbf{Value} \\ 
\midrule
Batch Size & 128 \\ 
Learning Rate & 5e-5 \\ 
Training Epoch & 4 \\ 
Learning Rate Deacy & Cosine \\ 
Warmup Step Ratio & 0 \\ 
Weight Decay & 0.1 \\ 
Adam $\beta_1$ & 0.9 \\ 
Adam $\beta_2$ & 0.999 \\ 
\bottomrule
\end{tabular}
\caption{
Hyperparameters for \textsc{KnowDist}'s optimization for Llama-3.2-3.2B-Instruct.
}
\label{tab:know-hp-llama-3b}
\end{table}

\begin{table}[!t]
\centering
\fontsize{8.5pt}{0.82\baselineskip}\selectfont
\begin{tabular}{lc} 
\toprule
\textbf{Hyper-parameter} & \textbf{Value} \\ 
\midrule
Batch Size & 128 \\ 
Learning Rate & 2e-5 \\ 
Training Epoch & 4 \\ 
Learning Rate Deacy & Cosine \\ 
Warmup Step Ratio & 0 \\ 
Weight Decay & 0.1 \\ 
Adam $\beta_1$ & 0.9 \\ 
Adam $\beta_2$ & 0.999 \\ 
\bottomrule
\end{tabular}
\caption{
Hyperparameters for \textsc{ICLDist}'s optimization for Llama-3.2-3.2B-Instruct.
}
\label{tab:icl-hp-llama-3b}
\end{table}

\section{Futher Implementation Details}
\label{app:c}
\subsection{Hyperparameter Settings of Distillation}
\label{app:c.1}

The detailed hyperparameters for the two-stage optimization are provided for each model: Llama-3.2-1.2B-Instruct (Tables \ref{tab:know-hp-llama-1b} and \ref{tab:icl-hp-llama-1b}), Qwen-2.5-1.5B-Instruct (Tables \ref{tab:know-hp-qwen-1.5b} and \ref{tab:icl-hp-qwen-1.5b}), and Llama-3.2-3.2B-Instruct (Tables \ref{tab:know-hp-llama-3b} and \ref{tab:icl-hp-llama-3b}).

\subsection{Computational Cost of Distillation}

This section details the computational cost incurred during the construction of the distillation corpus, which totals 1.5 million samples across the \textsc{KnowDist} and \textsc{ICLDist} stages. We use Llama-3-70B as the teacher model for data generation, which incurs approximately 671 A100 GPU hours in total. Specifically, we employ LMDeploy\footnote{\url{https://github.com/InternLM/lmdeploy}} with 4$\times$A100 GPUs (40GB each), achieving an average generation speed of about 9k samples per hour with a batch size of 200.

\section{Additional Experimental Results}
\label{app:d}

\subsection{Evaluation of Teacher Quality}

The effectiveness of knowledge distillation is fundamentally dependent on teacher model quality. To evaluate this critical factor, we conduct a quantitative evaluation of our primary teacher model, Llama3-70B, focusing on its response quality and the potential noise in the distillation data. We include GPT-3.5 as a comparative baseline and employ Claude-4 as an automated evaluator to approximate human judgment in our evaluation.

\vspace{5pt}
\noindent
\textbf{Tasks.} We selecte a range of tasks from both the \textsc{ICLDist} and \textsc{KnowDist} stages to ensure a comprehensive evaluation. For the \textsc{ICLDist} stage, we include Sentiment Classification and Emotion Detection tasks. For the \textsc{KnowDist} stage, we evaluate response quality using our multi-perspective prompting method, which encompasses expression-perspective sentiment analysis, target-perspective analysis, emotion-perspective analysis, and background-perspective analysis.

\vspace{5pt}
\noindent
\textbf{Metrics.} For tasks in \textsc{ICLDist} stage, we randomly sample 50 examples per task and apply task-specific performance measures: accuracy for Sentiment Classification and F1 score for Emotion Detection. For tasks in \textsc{KnowDist} stage, we conduct a comprehensive evaluation by randomly sampling 20 responses per task and assessing them across five quality dimensions: result accuracy, result completeness, explanation accuracy, explanation completeness, and hallucination. Each dimension is scored on a 0-5 scale, with 5 representing a completely correct response.

\begin{table*}[t]
\centering
\fontsize{8.5pt}{0.83\baselineskip}\selectfont
\setlength\tabcolsep{2.23pt}
\begin{tabular}{llccccc}
\toprule
\textbf{Perspective} & \textbf{Model} & \textbf{Result Acc.} & \textbf{Result Comp.} & \textbf{Explanation Comp.} & \textbf{Explanation Acc.} & \textbf{Hall.} \\
\midrule
\multirow{2}{*}{Expression} & Llama3-70B & 4.25 & 4.55 & 4.60 & 4.25 & 5.00 \\
                           & GPT-3.5 & 3.45 & 3.80 & 4.00 & 3.70 & 4.80 \\
\midrule
\multirow{2}{*}{Target} & Llama3-70B & 4.60 & 4.75 & 4.75 & 4.65 & 4.65 \\
                       & GPT-3.5 & 4.40 & 4.75 & 4.75 & 4.40 & 4.75 \\
\midrule
\multirow{2}{*}{Emotion} & Llama3-70B & 4.55 & 4.55 & 4.65 & 4.60 & 4.70 \\
                        & GPT-3.5 & 4.55 & 4.60 & 4.50 & 4.60 & 4.70 \\
\midrule
\multirow{2}{*}{Background} & Llama3-70B & 4.45 & 4.55 & 4.65 & 4.55 & 4.60 \\
                           & GPT-3.5 & 4.30 & 4.50 & 4.45 & 4.60 & 4.70 \\
\bottomrule

\end{tabular}
\caption{Performance of Llama3-70B versus GPT-3.5 on the \textsc{KnowDist} dataset. Acc: Accuracy, Comp: Completeness, Hall: Hallucination.}
\label{tab:knowdist-human-evaluation}
\end{table*}

\begin{table*}[ht]
\centering
\fontsize{8.5pt}{0.84\baselineskip}\selectfont
\setlength\tabcolsep{0.50pt}
\begin{tabular}{
p{115pt}
>{\centering\arraybackslash}m{23pt}
>{\centering\arraybackslash}m{23pt}
>{\centering\arraybackslash}m{23pt}
>{\centering\arraybackslash}m{23pt}
>{\centering\arraybackslash}m{23pt}
>{\centering\arraybackslash}m{23pt}
>{\centering\arraybackslash}m{23pt}
>{\centering\arraybackslash}m{23pt}
>{\centering\arraybackslash}m{23pt}
>{\centering\arraybackslash}m{23pt}
>{\centering\arraybackslash}m{23pt}
>{\centering\arraybackslash}m{23pt}
>{\centering\arraybackslash}m{40pt}} 
\toprule
\multirow{2}*{\textbf{Models}} & \multicolumn{4}{c}{\textbf{BSA}} & \multicolumn{4}{c}{\textbf{MSA}} & \multicolumn{4}{c}{\textbf{FSA}} & \multirow{2}*{\textbf{Avg}}\\
\cmidrule(lr){2-5}  \cmidrule(lr){6-9} \cmidrule(lr){10-13} 
& {IMDb} & {Yelp2} & {SST2} 
& {Twitter} & {Irony} & {Emoti.}  
& {Stance} & {Intim.} & {ATSA} & {ACSA} & {ASQP} & SSA \\
\midrule

Qwen2.5-Math-1.5B-Instruct & 66.15 & 69.20 & 56.82 & 25.87 & 52.08 & 32.06 & 48.24 & 33.33 & 0 & 0 & 0 & 0 & 31.98\\

DeepSeek-R1-Distill-Qwen-1.5B & 77.87 & 87.91 & 80.41 & 56.23 & 58.38 & 49.78 & 60.88 & 35.04 & 28.92 & 29.83 & 1.12 & 4.79 & 47.60\sixf{15.62} \\
DeepScaleR-1.5B-Preview & 77.85 & 89.83 & 81.24 & 55.06 & 60.75 & 49.51 & 56.85 & 36.51 & 30.09 & 34.20 & 1.98 & 3.49 & 48.11\sixf{16.13}\\

\textbf{\textsc{Know} \& \textsc{ICLDist} (\textsc{ours})} & 90.79 & 95.70 & 86.35 & 62.95 & 66.34 & 70.61 & 64.42 & 32.76 & 18.34 & 26.34 & 7.46 & 13.13 & 52.93\sixf{20.95}\\

\bottomrule
\end{tabular}
\caption{
Performance comparison between our domain-specific distillation method and reasoning-enhanced baselines.
}
\label{tab:comprision-to-rl}
\end{table*}

\vspace{5pt}
\noindent
\textbf{Results.} Table \ref{tab:icldist-human-evaluation} and \ref{tab:knowdist-human-evaluation} suggest that Llama3-70B consistently outperforms GPT-3.5 across most tasks, indicating its superior quality as a teacher model. Specifically, Llama3-70B demonstrates strong performance in complex analytical tasks, with notable superiority in Target-perspective analysis where it achieves higher scores across multiple dimensions including result accuracy (4.60 vs 4.40) and explanation accuracy (4.65 vs 4.40). However, despite these promising results, it is important to acknowledge that even minor inaccuracies in the teacher model may adversely affect the student model's performance, highlighting the continued importance of teacher quality in knowledge distillation processes.

\begin{table}[t]
\centering
\fontsize{8.5pt}{0.83\baselineskip}\selectfont
\setlength\tabcolsep{2.23pt}
\begin{tabular}{l c ccc c ccc c ccc c ccc c c} 
\toprule

\textbf{Task} & Llama3-70B & GPT-3.5 \\
\midrule
Sentiment Classification & 92.00 & 84.00 \\
Emotion Detection  & 82.90 & 75.49 \\

\bottomrule
\end{tabular}
\caption{Performance of Llama3-70B versus GPT-3.5 on the \textsc{ICLDist} tasks. The metric is accuracy for Sentiment Classification and F1 score for Emotion Detection.}
\label{tab:icldist-human-evaluation}
\end{table}

\subsection{Comparison with Reasoning-Enhanced Methods}

Reasoning-enhanced methods \cite{openai2024openaio1card, deepseekai2025deepseekr1incentivizingreasoningcapability} have recently attracted significant attention. They use reinforcement learning or long chain-of-thought distillation to improve model reasoning capabilities, achieving impressive results on small language models, particularly in mathematical tasks. Therefore, we explore whether these methods can improve similarly performance in sentiment analysis and compare them with our domain-specific distillation approach.

\vspace{5pt}
\noindent
\textbf{Experimental Setup.}  
We apply our domain-specific distillation method to the Qwen2.5-Math-1.5B-Instruct\footnote{\url{https://huggingface.co/Qwen/Qwen2.5-Math-1.5B-Instruct}} model and select two representative models as baselines. The first, DeepSeek-R1-Distill-Qwen-1.5B\footnote{\url{https://huggingface.co/deepseek-ai/DeepSeek-R1-Distill-Qwen-1.5B}}, is enhanced through long CoT distillation. The second, DeepScaleR-1.5B-Preview\footnote{\url{https://huggingface.co/agentica-org/DeepScaleR-1.5B-Preview}}, leverages both long CoT distillation and reinforcement learning.

\vspace{5pt}
\noindent
\textbf{Results.}
The results in Table \ref{tab:comprision-to-rl} reveal that reasoning-enhanced models achieve improvements in sentiment analysis tasks. Our domain-specific distillation method achieves even more significant performance gains, demonstrating its superior effectiveness for sentiment analysis tasks. These findings suggest that combining both approaches—reasoning enhanced and domain-specific distillation—presents a promising direction for future research, potentially unlocking further performance improvements beyond what either method achieves individually.

\begin{figure*}[b]
\centering
\includegraphics[width=0.95\linewidth]{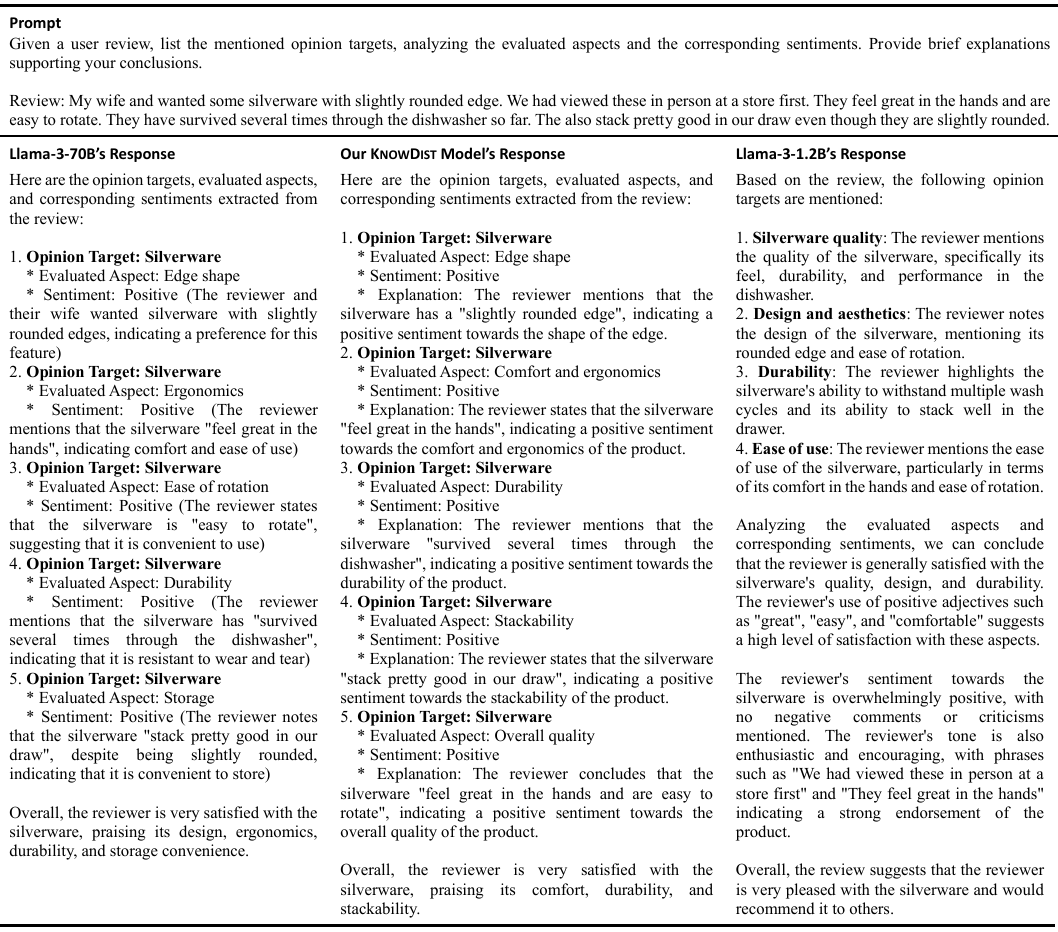}
\caption{
Representative example for sentiment analysis.
}
\label{fig:case-1}
\end{figure*}

\begin{figure*}[h]
\centering
\includegraphics[width=0.95\linewidth]{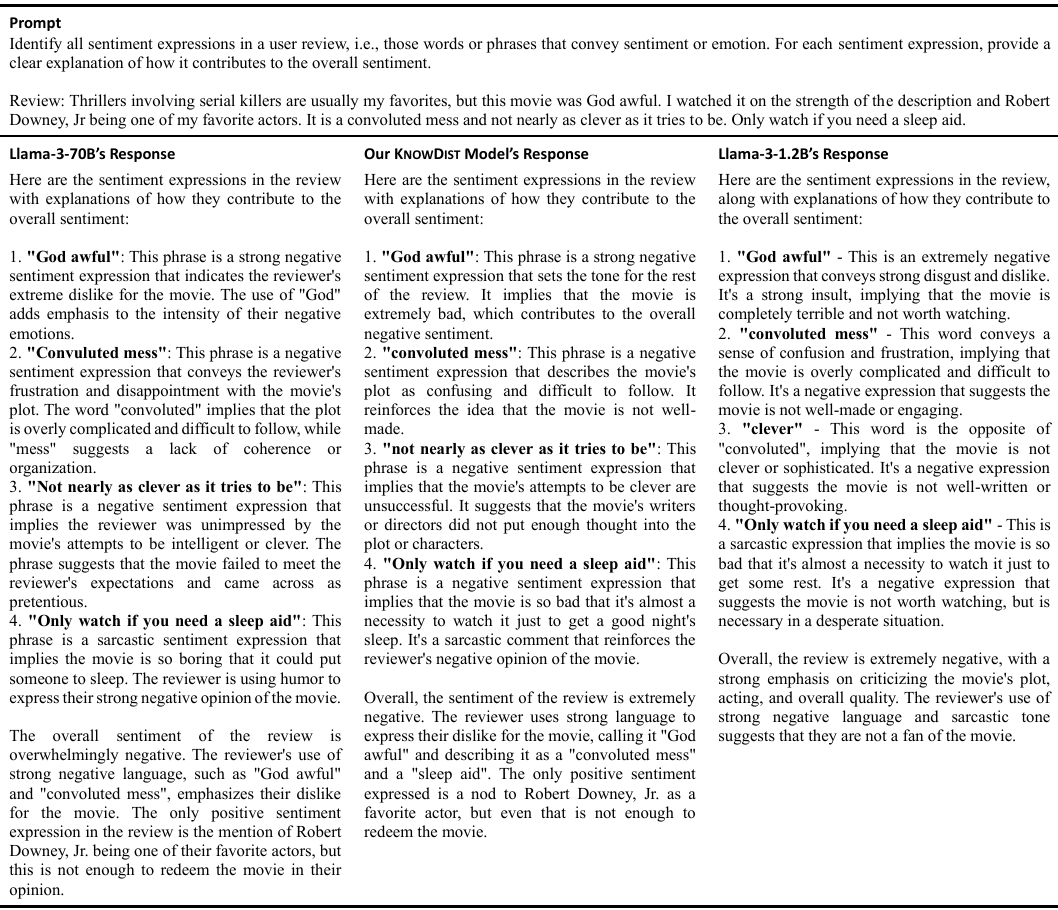}
\caption{
Representative example for sentiment analysis.
}
\label{fig:case-2}
\end{figure*}

\subsection{Case Study}
Figures \ref{fig:case-1} and \ref{fig:case-2} present representative examples to demonstrate the basic sentiment analysis capabilities of Llama-3-70B, our model, and Llama-3-1.2B. Among the three models, Llama-3-70B achieves the best analysis results, followed by our model, while Llama-3-1.2B shows the weakest performance. The key differences are reflected in three aspects: the accuracy and comprehensiveness of the analysis results, as well as the depth of reasoning.

\end{document}